\pdfoutput=1

\documentclass[11pt]{article}

\usepackage[]{EMNLP2022}

\usepackage{times}
\usepackage{latexsym}

\usepackage[T1]{fontenc}

\usepackage[utf8]{inputenc}

\usepackage{microtype}

\usepackage{times}
\usepackage{latexsym}

\usepackage{microtype}
\usepackage{todonotes}
\usepackage{url}
\usepackage{subfiles}
\usepackage{wrapfig}
\usepackage{float}
\usepackage{caption}
\usepackage{tikz}
\usetikzlibrary{decorations.pathreplacing}
\usepackage{graphicx}
\usepackage{subcaption}
\usepackage{booktabs}

\usepackage[utf8]{inputenc} 
\usepackage[T1]{fontenc}    
\usepackage{hyperref}       
\usepackage{amsfonts}       
\usepackage{nicefrac}       
\usepackage{pifont}
\usepackage{multirow}
\usepackage{paralist}
\usepackage{xspace}
\usepackage{tcolorbox}
\usepackage{multirow}
\usepackage{diagbox}
\usepackage{inconsolata} 
\usepackage{colortbl}
\usepackage{xcolor}

\usepackage{bbding}
\usepackage{pifont}
\usepackage{wasysym}
\usepackage{amssymb}
\usepackage{makecell}

\newcommand\sect[1]{\S\ref{#1}}

\newcommand{\vcr}{\textsc{VCR}\xspace}
\newcommand{\vqax}{\textsc{VQA-X}\xspace}
\newcommand{\esnlive}{\textsc{E-SNLI-VE}\xspace}
\newcommand{\vlp}{\textsc{VLP}\xspace}

\newcommand{\vatfive}{\textsc{VA-T5}\xspace}

\newcommand{\clip}{\textsc{CLIP}\xspace}

\definecolor{myred}{RGB}{224,34,20}
\definecolor{mygreen}{RGB}{60,143,43}

\title{On Advances in Text Generation from Images Beyond Captioning:\\A Case Study in Self-Rationalization} 

\author{\begin{tabular}{c}
    Shruti Palaskar$^{\dagger\clubsuit*}$ \quad
    Akshita Bhagia$^\ddag$ \quad 
    Yonatan Bisk$^\dagger$ \vspace{.5mm} \\
    Florian Metze$^\dagger$ \quad
    Alan W Black$^\dagger$ \quad
	Ana Marasovi\'{c}$^{\S*}$
	\end{tabular}
	\\ \vspace{.5mm}
	\begin{tabular}{c}
	\\ \vspace{.5mm}
	$^\dagger$Carnegie Mellon University, Pittsburgh, PA, USA \\
	$^\ddag$Allen Institute for AI, Seattle, WA, USA \\
	$^\S$University of Utah, Salt Lake City, UT, USA \\
	\end{tabular}
	\\ \vspace{.5mm}
    \small
    \begin{tabular}{c}
\texttt{\{spalaska,ybisk,fmetze,awb\}@cs.cmu.edu} \quad  \texttt{akshitab@allenai.org} \quad \texttt{ana.marasovic@utah.edu}
\end{tabular}
}
\vspace{5mm}

\begin{document}
\maketitle
\begingroup\def\thefootnote{*}\footnotetext{Work undertaken while Shruti Palaskar and Ana Marasovi\'{c}  were at the Allen Institute for AI.}\endgroup
\begingroup\def\thefootnote{$\clubsuit$}\footnotetext{Shruti Palaskar is currently at Apple.}\endgroup

\begin{abstract}
Combining the visual modality with pretrained language models has been surprisingly effective for simple descriptive tasks such as image captioning. More general text generation however remains elusive. %
We take a step back and ask: How do these models work for more complex generative tasks, i.e.~conditioning on both text and images? %
Are multimodal models simply visually adapted language models, or do they combine they reason jointly over modalities?

We investigate these questions in the context of \emph{self-rationalization} (jointly generating task labels/answers and free-text explanations) of three tasks: (i) visual question answering in \vqax, (ii) visual commonsense reasoning in \vcr, and (iii) visual-textual entailment in \esnlive. %
We show that recent unimodal advances, CLIP image representations and scaling of language models, do not consistently improve self-rationalization in multimodal tasks. %
We find that no single model type works universally best across tasks, datasets, and finetuning data sizes. %
Our ﬁndings motivate the need for novel general backbones approach 
that move text generation from images and text beyond image captioning. 
\end{abstract}


\section{Introduction} 
\begin{figure*}
    \centering
    \includegraphics[width=\textwidth]{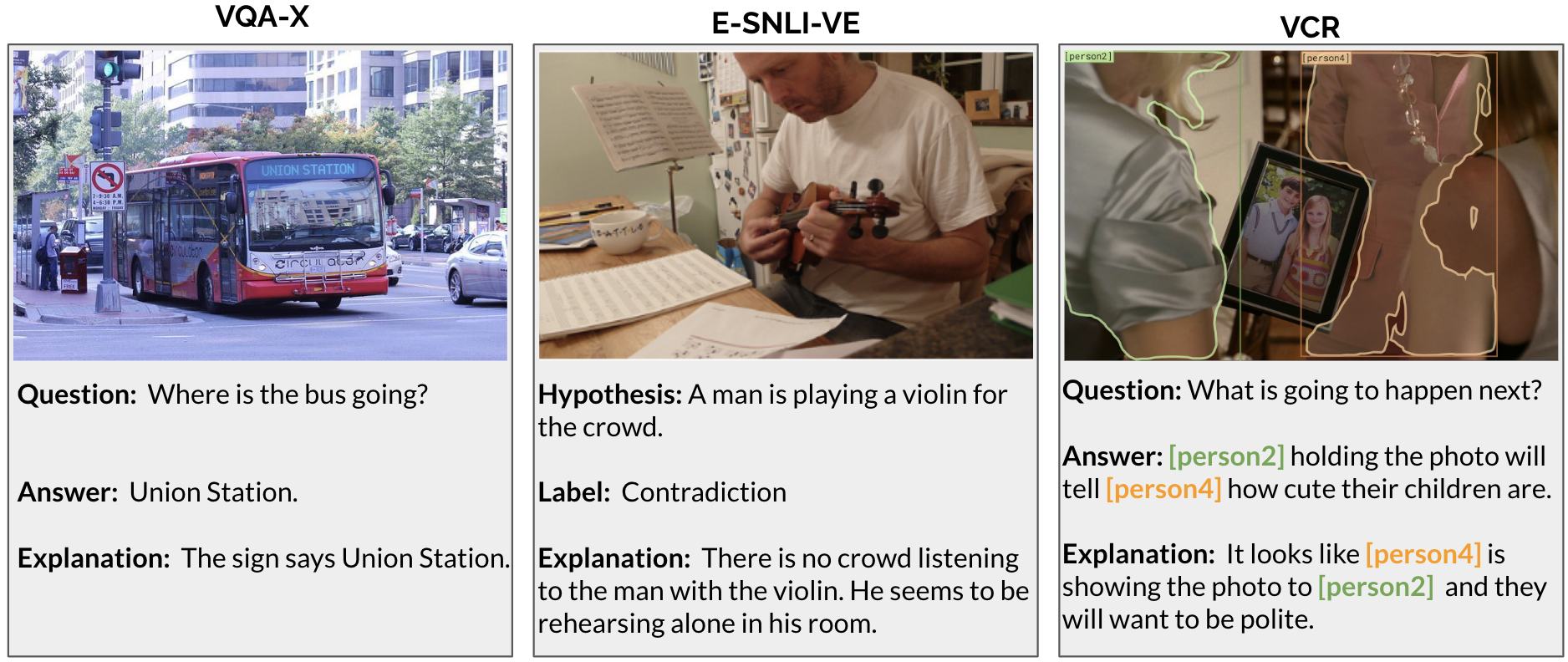}
    \caption{Self-rationalization tasks.}
    \label{fig:illustration}
\end{figure*}

The pretrain-finetune paradigm has changed the field of NLP. %
Inspired by its success, there has been an explosion of interest in multimodal pretraining \cite{Su2019VLBERTPO, Lu2019ViLBERTPT,Chen2019UNITERLU,Li2019UnicoderVLAU, tan-bansal-2019-lxmert,li2020oscar,Gui2022}. %
To enable text generation from images, captioning is often included as one of the pretraining tasks \cite{zhou2020unified,gupta2021towards,wang2022SimVLM}. %
Captioning is also the only generative task used to evaluate and compare joint models, for which only minor improvements are reported 
relative to classification tasks \cite{pmlr-v139-cho21a,Chang2022}. %
Moreover, captioning is conditioned only on a single image. %
This leads us to ask: \emph{Do recent advances transfer to more complex generative tasks? %
Can generation condition on both images and text?} 
Another line of work skips joint pretraining and directly modifies and finetunes a pretrained language model apt for generation \cite[e.g., GPT-2;][]{Radford2019LanguageMA} on multimodal datasets \cite{Park2020VisualCG,Sollami2021MultimodalCF,Eichenberg2021MAGMAM,Gui2022KAT}. %
This approach has distinct benefits and downsides compared to models based on joint  pretraining (see Table \ref{tab:benefits_downsides}), but these two families of models are rarely compared.
This leads us to other questions: \emph{Given a new generative task, which  approaches should be used or combined?} 

We study these questions through the lens of a newly emerging and important, but challenging task of \emph{self-rationalization} \cite{wiegreffe-etal-2021-measuring}: jointly generating both the task label/answer and a free-text explanation for the prediction. %
Standard tasks for studying multimodal self-rationalization present different levels of difficulty. %
Explaining \textsc{VQA} is similar to captioning since corresponding explanations describe  \emph{visible} information in the image that is relevant to the context (Fig.\ \ref{fig:illustration}, left). %
On the other hand, \vcr instances require higher-order reasoning about \emph{unstated}  information such as commonsense (Fig.\ \ref{fig:illustration}, right). %
Since \vcr answers are full sentences, self-rationalization consists of two generative sub-tasks. %

We evaluate the following models: (i) a joint vision-language model, \vlp \cite{zhou2020unified}, (ii) a pretrained language model, T5 \cite{raffel2019exploring}, that we visually adapt only through finetuning, and (iii) VL-T5/VL-BART \cite{pmlr-v139-cho21a}, a combination of the previous two approaches. %
Namely, VL-T5/VL-BART are developed from T5 and BART \cite{lewis-etal-2020-bart} by doing a second round of pretraining, using multimodal datasets and objectives. We finetune all models for self-rationalization in two data settings: (i) using entire finetuning datasets (high-resource setting), and (ii) using 30\% of the data (low-resource setting). %

We first present an analysis of the factors influencing performance of the visually adapted T5: the choice of image representation (\sect{subsec:va_t5_features}) and T5's model size (\sect{sec:va_t5_model_size}). %
We demonstrate that recent advances in representing images, namely CLIP features, can be easily leveraged to get more accurate visually adapted T5 models. %
However, these improvements are not realized for a more complex sub-task of explanation generation. %
Moreover, unlike for text generation conditioned only on text (including self-rationalization of tasks with textual inputs; \citealt{marasovic2022feb}) we do not see clear performance improvements from scaling visually adapted PLMs. %
Finally, the main comparison of the three model types described above  (\sect{sec:main_comparison}) shows that no model type works universally the best across tasks and data regimes. %
These results demonstrate that outside of image captioning, there is no clear choice of model backbone or training regime in image conditioned text generation.
We aim to motivate research on multimodal model comparisons across generative tasks and experimental setups, to help realize  benefits of different model families (Table \ref{tab:benefits_downsides}) with a single approach.  
\begin{table*}[t]
\small
\centering
\resizebox{\textwidth}{!}{  
\begin{tabular}{ll|p{1cm}p{1cm}p{1cm}}
\toprule
&\multirow{4}{*}{\textbf{Wanted Model Properties for Multimodal Self-Rationalization}} & \textbf{Visually Adapted PLMs} (\sect{subsec:vision_lms}) & \textbf{Joint VL Models} (\sect{subsec:unified_models}) & \textbf{Comb. Models} (\sect{subsec:combined approaches})\\
\midrule
1& Designed for some text generation task (e.g., image captioning, language modeling) & \makecell[c]{\ding{51}} & \makecell[c]{Some} & \makecell[c]{Some} \\
2 & Offered in larger sizes (related to better text generation performance) & \makecell[c]{\ding{51}} & & \\
3 & Large textual pretraining data (related to capturing world/commonsense knowledge) & \makecell[c]{\ding{51}} & & \makecell[c]{\ding{51}}\\
4 & Easy plug-and-playing with the latest pretrained LMs and image representations & \makecell[c]{\ding{51}} & & \\
5 & Tight coupling between modalities & & \makecell[c]{\ding{51}} & \makecell[c]{\ding{51}} \\
6 & Expected to be robust when multimodal training data is limited & & \makecell[c]{\ding{51}} & \makecell[c]{\ding{51}} \\
\bottomrule
\end{tabular}}
\caption{A comparison between: training vision and language (VL) jointly from scratch, adapting pretrained language models (PLM) to visual features, and models that somewhat combine these two approaches, w.r.t.\ desired model properties for self-rationalization. Text generation typically improves with model size \cite{brown2020gpt3}, incl.\ self-rationalization \cite{marasovic2022feb}. Due to huge pretraining corpora, PLMs have been shown to capture some world \cite{petroni-etal-2019-language} and commonsense knowledge \cite{davison-etal-2019-commonsense} which is beneficial for self-rationalization as the task often requires \emph{inferring} relevant information from the inputs (see Fig.\ \ref{fig:illustration}).}
\label{tab:benefits_downsides}
\end{table*}


\begin{table*}[t]
\small
\centering
\resizebox{\textwidth}{!}{  
\begin{tabular}{lp{2.5cm}p{2.75cm}p{2.6cm}p{2.6cm}}
\toprule
\textbf{Model} & \textbf{Backbone} & \textbf{\makecell[tl]{Backbone PT\\Objectives}} & \textbf{Backbone PT Data} & \textbf{\makecell[tl]{Continued PT\\Datasets}} \\
\midrule
VLP            & BERT / UniLM & MLM                                                                               & Wiki, BookCorpus                                                                         & Conceptual Captions                   \\
\midrule
VA-T5-CLIP     & \multirow{4}{*}{T5}             & \multirow{4}{3cm}{Fill-in-the-span-style denoising objectives + multitask learning} & \multirow{4}{3cm}{C4 + a suite of annotated datasets for classification, QA, translation, etc.} & \multirow{3}{3cm}{\emph{None}}        \\
VA-T5-Captions & &  &    &                                    \\
VA-T5-Objects  &  &            &                                                                                         &                         \\
\cmidrule{1-1}\cmidrule{5-5}
VL-T5 & & & &   \multirow{4}{3cm}{MS COCO, Visual Genome, VQA v2.0, GQA, Visual7W}                                         \\
\cmidrule{1-4} 
VL-BART        & BART                            & Reconstruct text corrupted with an
arbitrary noising function                                    &   Wiki, BookCorpus, Stories, CCNews, OpenWebText                                                                                              &                               \\
\toprule
\textbf{Model}          &  \textbf{\makecell[tl]{Continued PT\\Objectives}} & \textbf{Img.\ Feat.\ Model}          & \textbf{\makecell[tl]{Img.\ Feat.\ Model \\ Backbone}} & \textbf{\makecell[tl]{Img.\ Feat.\ Model\\PT Data}} \\
\midrule 
VLP            & MLM, LM  & Faster R-CNN  & ResNeXt-101 FPN & Visual Genome \\
\midrule 
VA-T5-CLIP     & \multirow{3}{3cm}{\emph{None}}  & CLIP & Vision Transformer &  New unavail.\ data\\
VA-T5-Captions &      & VL-T5  &  See VL-T5 rows    &   See VL-T5 rows   \\
VA-T5-Objects  &     &  Faster R-CNN  &  ResNeXt-101 FPN & Visual Genome\\
\midrule 
VL-T5          &  \multirow{2}{3cm}{MLM, VQA, image-text matching, visual grounding, grounded captioning}  & \multirow{2}{2cm}{Faster R-CNN }     &   \multirow{2}{3cm}{ResNeXt-101 FPN}                             &   \multirow{2}{3cm}{Visual Genome}                                                                                                                                                                                                                   \\
VL-BART        &    &                              &                                                                                                                                                                                                                 \\
&    &                              &                                                                                                                                                                                                                 \\
&    &                              &                                                                                                                                                                                                                 \\

\bottomrule
\end{tabular}
}
\caption{Model specifications. PT stands for ``pretraining'', MLM for ``masked language modeling'', ``Img.\ Feat.'' for ``Image Features''.  Sources: BERT \cite{devlin-etal-2019-bert}, UniLM \cite{DBLP:conf/nips/00040WWLWGZH19}, BookCorpus \cite{DBLP:conf/iccv/ZhuKZSUTF15}, T5 \cite{raffel2019exploring}, C4 \cite{raffel2019exploring} is made publicly available by \citet{dodge-etal-2021-documenting}, BART \cite{lewis-etal-2020-bart}, MS COCO \cite{lin2014microsoft}, VQA v2 \cite{goyal2017making}, GQA \cite{hudson2019gqa}, Visual7W \cite{zhu2016visual7w}, Stories \cite{DBLP:journals/corr/abs-1806-02847}, \href{https://commoncrawl.org/2016/10/news-dataset-available/}{CCNews}, OpenWebText \cite{Gokaslan2019OpenWeb}, Conceptual Captions \cite{sharma2018conceptual}, Faster R-CNN \cite{Ren2015FasterRT}, Visual Genome \cite{krishna2017visual}, ResNeXt-101 FPN \cite{DBLP:conf/cvpr/XieGDTH17}, Vision Transformer \cite{DBLP:conf/iclr/DosovitskiyB0WZ21}.}
\label{tab:overview}
\end{table*}

\section{Text Generation from Images: Models}
\label{sec:models_background}
Vision-and-language (VL) learning currently comprises two families of models: 
(i) \emph{joint VL models} that are pretrained from scratch using data with both modalities (\sect{subsec:unified_models}), and (ii) \emph{vision-adapted language models}---pretrained language models adapted to the visual modality through finetuning using end-task multimodal datasets (\sect{subsec:vision_lms}). %
Some models combine these two approaches to some extent, so they could be the best of both worlds (\sect{subsec:combined approaches}). %

In Table \ref{tab:benefits_downsides}, we overview reasons for why one model family might be preferred over the other. %
In Tables \ref{tab:overview}--\ref{tab:img_sources}, we outline model specifications and list image sources. %
For generative tasks conditioned on images, including self-rationalization of VL tasks, the choice of the best base model family is not obvious. %
The aim of this paper is to find whether such a choice exists. 


\subsection{VLP: Joint Vision-and-Language Model} 
\label{subsec:unified_models}

We use VLP \cite{zhou2020unified} to analyze importance of benefits of joint VL pretraining from scratch relative to other approaches.

VLP is a shared encoder-decoder transformer \cite{vaswani2017attention} of size similar to BERT-base \cite[110M parameters;][]{devlin-etal-2019-bert}. %
It is pretrained with objectives similar to both masked and standard language modeling. %
Thus, it is suitable for discriminative as well as generative tasks. %
A given input image is represented with vector representations of a fixed number of regions obtained with an off-the-shelf object detector \cite{wu2019detectron2}. %
During finetuning, the same object detector representations should be used. %
The Conceptual Captions dataset \cite{sharma2018conceptual}, containing about 3M web-accessible images and associated captions, is used for pretraining. 


\subsection{VA-T5: Vision-Adapted Pretrained LM}
\label{subsec:vision_lms}

Vision-adapted training involves starting with a pretrained language model (PLM) and adapting it to VL tasks as a finetuning step.  We start with T5 \cite{raffel2019exploring}---a PLM that is commonly used for self-rationalization \cite{DBLP:journals/corr/abs-2004-14546, hase-etal-2020-leakage, wiegreffe-etal-2021-measuring, marasovic2022feb}---and finetune it for self-rationalization of VL tasks. %
Specifically, we concatenate image representations with representations of textual inputs, feed the result 
to the subsequent PLM's  layers, and train 
using language modeling loss on generated answer and explanation tokens. %

A question that emerges is: what kind of image representations should be used? %
While vector representations of image regions extracted with an object detector are the most frequent choice, most recently advantages of representations from the CLIP model \cite{pmlr-v139-radford21a} have been demonstrated for various applications \cite{Shen2021HowMC}. %
Moreover, we wonder whether using automatic image captions is the way to visually adapt a PLM given that the input will then be completely textual, i.e, in the modality a PLM has seen before. %
Thus, in \sect{subsec:va_t5_features}, we compare three different features: (i) auto-generated captions from the off-the-shelf image captioning model \cite[VL-T5 model;][]{pmlr-v139-cho21a}, (ii) object features from a pre-trained R-CNN model \cite{Ren2015FasterRT}, and (iii) CLIP features (obtained from the last layer of ViT-B/32).
Besides exploring different image representations, visually adapting a PLM allows us to study model scaling. %
In \sect{sec:va_t5_model_size}, we study the following sizes of T5: Base (220M parameters), Large (770M), and 3B. 
We refer to visually adapted T5 as VA-T5. 


\subsection{VL-T5 / VL-BART: Combined Models}
\label{subsec:combined approaches}

Another approach is to start with a PLM, do a second round of pretraining to learn joint VL representations, and finally finetune the model to learn the end-task. %
This approach can be seen both as a joint model and a visually adapted PLM, and thus may offer benefits from the both model families.  

To compare this approach with the others, we use VL-T5 and VL-BART \cite{pmlr-v139-cho21a}. %
VL-BART, a multimodal extension of BART-Base \cite[139M parameters;][]{lewis-etal-2020-bart}, also follows an encoder-decoder transformer but does not share the parameters between the encoder and decoder as is done in VLP. 
To represent images, it uses the Faster R-CNN model for object detection \cite{Ren2015FasterRT}. %
Specifically, vector representations of a fixed number of image regions are are concatenated with text embeddings and fed into VL-BART, which is then pretrained using masked language modeling objectives in addition to new objectives such as visual question answering, image-text matching, visual grounding, and grounded captioning. %
VL-T5 is similar in spirit to VL-BART, where it is initialized with a T5-Base model \cite[220M parametes;][]{raffel2019exploring} correspondingly. %
T5 is trained for various downstream tasks jointly, whereas BART exploits a task-specific encoder-decoder set up for sequence generation tasks. %
Both VL-BART and VL-T5 are pre-trained with MS COCO \cite{lin2014microsoft}, Visual Genome \cite{krishna2017visual}, VQA v2 \cite{goyal2017making}, GQA \cite{hudson2019gqa}, and Visual7W \cite{zhu2016visual7w}, leading to a total of 9.18M image-text pairs on 180K unique images.


\section{Experimental Setup}
\label{sec:exp_setup}
In this section, we describe tasks, datasets, and evaluation setup for self-rationalization of vision-and-language tasks introduced in prior work \cite{marasovic-etal-2020-natural, Kayser2021eViLAD}.


\subsection{Tasks and Datasets} 

The dataset statistics are given in Table \ref{tab:dataset_stats}, where the average answer and explanation lengths hint on differences in complexities of each task. The three datasets represent different levels of required reasoning (see examples in Figure \ref{fig:illustration}), e.g., are the images representing simpler scenarios (Flickr30K) or complex movie scenes (VCR).

\begin{table}[!ht]
\small
\centering
\resizebox{\columnwidth}{!}{  
\begin{tabular}{ll}
\toprule
\textbf{Dataset}       & \textbf{Image Sources}            \\
\midrule
VQA-X         & VQA v2.0                  \\
E-SNLI-VE     & SNLI / \href{https://www.flickr.com/}{Flickr}               \\
VCR           & movie clips / \href{https://www.youtube.com/user/movieclips}{Fandango}    \\
\midrule
MS COCO       & \href{https://www.flickr.com/}{Flickr}                    \\
YFCC100M      & \href{https://www.flickr.com/}{Flickr}                    \\
Visual Genome & YFCC100M + MS COCO        \\
VQA v2.0      & MS COCO                   \\
GQA           & Visual Genome             \\
Visual7W      & MS COCO \\
\bottomrule
\end{tabular}
}
\caption{Image sources. MS COCO \cite{lin2014microsoft}, SNLI \cite{bowman-etal-2015-large},  movie clips \cite{DBLP:journals/ijcv/RohrbachTRTPLCS17}, YFCC100M \cite{DBLP:journals/cacm/ThomeeSFENPBL16}.}
\label{tab:img_sources}
\end{table}

\noindent \textbf{\vqax \cite{Park2018MultimodalEJ}} is the extension of the widely-used Visual Question Answering v1 \cite{antol2015vqa} and v2 \cite{goyal2017making} datasets, with corresponding free-text explanations. The images here are originally sourced from the MSCOCO dataset \cite{lin2014microsoft}, and the answers are collected for open-ended questions about these images that require vision, language, and commonsense knowledge to answer. 

\begin{table*}[t]
\centering
\resizebox{\textwidth}{!}{  
\begin{tabular}{llccc}
\toprule
\textbf{Dataset} & \textbf{Task}   & \textbf{\# Samples} & \textbf{Avg.\ Answer Len} & \textbf{Avg.\ Explanation Len}  \\ 
         &      & \small{train/val/test} & \small{train/val/test} & \small{train/val/test} \\
\midrule
\vcr   & Visual Commonsense Reasoning & 212.9K / 26.5K / 25.2K       &  7.54 / 7.65 / 7.55                 &    16.16 / 16.19 / 16.07 \\
\esnlive       & Visual Entailment  & 402K / 14K / 15K        &   1/1/1     &   12.3/13.3/13.2  \\
\vqax       & Visual Question Answering &  29.5K / 1.5K / 2K    &  1.03/1.05/1.03    &  8.6/9.0/9.2 \\
\bottomrule
\end{tabular}}
\caption{Specifications of the self-rationalization datasets.}
\label{tab:dataset_stats}
\end{table*}

\noindent \textbf{\esnlive \cite{Kayser2021eViLAD}} is a dataset for visual-textual entailment, the task of predicting whether a statement is entailed, contradicted, or neutral given an image that serves as a premise. %
\esnlive combines annotations from two datasets: (i) SNLI-VE \cite{xie2019visual}, collected by replacing the textual premises of SNLI \cite{bowman-etal-2015-large} with Flickr30K images \cite{young-etal-2014-image}, and (ii) E-SNLI \cite{camburu2018snli}, a dataset of crowdsourced free-text explanations for SNLI. %

\noindent \textbf{\vcr \cite{Zellers2019FromRT}} is a carefully crowdsourced dataset of answers and explanations for visual scenes extracted from Hollywood movies. %
Thus, the visual context in this data is more complex than MSCOCO or Flickr30K images, leading to more complex answers and explanations. %
\citeauthor{Zellers2019FromRT} instructed crowdworkers to first annotate answers for a given question-image pair, and then showed the annotated answer along with the question-image pair to a different set of annotators to get the corresponding explanation. %
This dataset was orginally introduced in a classification setting: given a question about an image, pick the correct answer from four choices, and then pick the correct explanation again from four choices. %
\citet{Dua2021BeyondVG} propose instead to generate both the answer and explanation. %
This is more realistic than the multiple-choice setting which is restricted to a user giving answer choices. %
In this paper, we also generate VCR answers and explanations.


\subsection{Evaluation Metrics} 
\label{subsec:eval_metrics}

Self-rationalization requires evaluating two sub-tasks: correctness of predicted answers/labels, and quality of generated explanations. %
For the former, we use \textbf{accuracy} for E-SNLI-VE and VQA-X. %
Evaluating VCR answers is more complicated as they are full sentences. %
Following \citet{Dua2021BeyondVG}, given a generated answer, we normalize text (remove articles, punctuation, lowercase), and count the number of overlapping words with the four available answer choices in the VCR dataset. %
We select an answer candidate with the highest overlap as the predicted answer. %
\textbf{Proxy accuracy} is the accuracy that is computed between the correct answer candidate and predicted answer candidate. 
\citet{Dua2021BeyondVG} do not report the correlation between proxy accuracy and human judgments of answer plausibility. %
To this end, for 600 VCR instances, we ask 5 crowdworkers to respond to ``Given the image and the question, is this answer likely?'' with yes, weak yes, weak no, or no, and map answers to 1, 2/3, 1/3, and 0, respectively. 
\textbf{Answer plausibility} is the average of scores of 5 annotators. 
In \sect{sec:results}, we report the average answer plausibility across 600 instances, as well as proxy accuracy.
Spearman's correlation coefficient between the proxy accuracy and answer plausibility is 0.56 (p < 0.028) indicating a moderate correlation between them.

Automatic metrics are unreliable, so human evaluation has been used to evaluate free-text explanation generation \cite{camburu2018snli,Kayser2021eViLAD,clinciu-etal-2021-study}. 
We ask 3 annotators whether an explanation justifies an answer/label given an image, questions/hypothesis, and a generated answer/label. %
Annotators 
pick one of the four options (yes, weak yes, weak no, no), and the four options are assigned numerical values (1, 2/3, 1/3, 0). %
We average scores of 3 annotators to get the plausibility of an individual explanation, and report the average \textbf{explanation plausibility} in a sample of 300 instances. %
Following \citet{Kayser2021eViLAD}, we select the first 300 instances for which the answer/label is correctly generated. %
For \esnlive, we select an equal number of examples for each label to produce a balanced evaluation set. 
Human evaluation used Amazon Mechanical Turk.\footnote{ %
Each batch of evaluation contains 10 samples. %
We pay \$1.5 per batch to each annotator for \vcr evaluations and \$1 per batch for \esnlive and \vqax each as \vcr tasks require longer time to complete.} %

\citet{Kayser2021eViLAD} report that all automated metrics are weakly correlated with explanation plausibility (per humans), but that BERTscore is most correlated. 
Therefore, we report \textbf{BERTscores} for completeness and reproducibility. %
Following \citet{Kayser2021eViLAD}, we set the BERTscore of incorrectly predicted instances to 0. 

\section{Results} 
\label{sec:results}
To study whether there is a base model family that is more suitable for text generation conditioned on images and text, %
we compare: (i) a joint vision-and-language model, VLP (\sect{subsec:unified_models}), (ii) a visually adapted PLM, VA-T5 (\sect{subsec:vision_lms}), and (iii) VL-T5 / VL-BART, a combination of the previous two model families (\sect{subsec:combined approaches}), for self-rationalization of the three tasks in Figure \ref{fig:illustration} and Table \ref{tab:dataset_stats}.
The benefits and downsides of (i) and (ii) are outlined in Table \ref{tab:benefits_downsides}.
Before presenting the outcomes of this comparison in \sect{sec:main_comparison}, 
we study the impact of the choice of image features and model size on VA-T5's performance.

\subsection{VA-T5: Analysis of Image Features}
\label{subsec:va_t5_features}

Visually adapting PLMs allows us to combine different image features with PLM's text representations. %
We analyze VA-T5-Base (finetuned with the full training data) with different image features in the input: auto-generated captions, object, and CLIP features (see \sect{subsec:vision_lms} for more information). %
We also report a control setting where no image features are used (\textit{None}). %
In Table \ref{tab:results_va_t5_features_accuracy}, we report \emph{answer} (proxy) accuracy and plausibility (for VCR), and in Tables \ref{tab:results_va_t5_features_bertscore} and \ref{tab:results_va_t5_features_plausibility} \emph{explanation} BERTscore and plausibility. %
Metrics are described in \sect{subsec:eval_metrics} and %
hyperparameters are reported 
in the Appendix. 

\begin{table}[t]
\footnotesize
\begin{subtable}{\columnwidth}
\resizebox{\columnwidth}{!}{  
\begin{tabular}{l|cccc}
\toprule
\textbf{Task} \textbackslash{} \textbf{Feat.} & None    & Captions & Objects& \clip \\
\midrule
\multirow{2}{*}{VCR}       & 54.5   & 56.2   & 57.8  & \textbf{58.1}\\
& 29.4 & 34.4 & 37.4 & \textbf{41.4} \\ 
\arrayrulecolor{black!20}\midrule
E-SNLI-VE & 67.6 & 71.7 & 72.5 & \textbf{74.7}  \\
VQA-X     & 43.4 & 73.8 & 72.3 & \textbf{74.7}\\
\arrayrulecolor{black}\bottomrule
\end{tabular}
}
\caption{Answer (Proxy) Accuracy / Plausibility$^\dagger$}
\label{tab:results_va_t5_features_accuracy}
\end{subtable}\par%
\vspace*{0.75em}
\begin{subtable}{\columnwidth}
\resizebox{\columnwidth}{!}{  
\begin{tabular}{l|cccc}
\toprule
\textbf{Task} \textbackslash{} \textbf{Feat.}  & None   & Captions & Objects & \clip \\
\midrule
VCR & 85.3 & 85.7 & 85.6 & \textbf{86.0} \\
E-SNLI-VE & 89.1 & \textbf{89.3} & \textbf{89.3} & \textbf{89.3} \\
VQA-X  & 90.7 & \textbf{91.2} & 91.1 & 90.9  \\
\bottomrule
\end{tabular}
}
\caption{Explanation BERTscore}
\label{tab:results_va_t5_features_bertscore}
\end{subtable}\par%
\vspace*{0.75em}
\begin{subtable}{\columnwidth}
\resizebox{\columnwidth}{!}{  
\begin{tabular}{l|cccc}
\toprule
\textbf{Task} \textbackslash{} \textbf{Feat.} & None   & Captions & Objects & \clip \\
\midrule
VCR & 15.6 & 19.6 & \textbf{21.0} & \textbf{21.0} \\
E-SNLI-VE & 65.4 & \textbf{66.6} & 65.7 & 65.2 \\
VQA-X & 70.8 & 75.5 & \textbf{76.4} & 75.8 \\
\bottomrule
\end{tabular}
}
\caption{Explanation Plausibility}
\label{tab:results_va_t5_features_plausibility}
\end{subtable}
\caption{A comparison of image features (Captions, Objects, CLIP) for \vatfive-Base (\sect{subsec:vision_lms}). $\dagger$ For VCR answers, we report both proxy accuracy (1st row) and  plausibility (2nd row; \sect{subsec:eval_metrics}).}
\label{tab:results_va_t5_features}
\end{table}
\begin{table*}
\small
\centering
\resizebox{0.95\textwidth}{!}{  
\begin{tabular}{l*{9}{>{\centering\arraybackslash}p{2.75em}}}
\toprule 
\multirow{2}{*}{\textbf{Task} \textbackslash{} \textbf{Model Size}} & \multicolumn{3}{c}{\textbf{Answer Accuracy$^\dagger$}} & \multicolumn{3}{c}{\textbf{Explanation BERTscore}} & \multicolumn{3}{c}{\textbf{Explanation Plausibility}} \\
\cmidrule(lr){2-4}\cmidrule(lr){5-7}\cmidrule(lr){8-10}
 & Base    & Large & 3B & Base    & Large & 3B & Base    & Large & 3B \\
\midrule
\multirow{2}{*}{VCR} & 58.1  & \textbf{59.1} & \textbf{59.1} & \textbf{86.0} & \textbf{86.0} & 85.8 & 21.0 & 24.4 & \textbf{25.8}
\\
 & 41.4 & \textbf{45.8} & 42.2 & - & -& -& -& -&-\\
 \arrayrulecolor{black!20}\midrule
E-SNLI-VE & \textbf{74.7} & 74.4 & 68.0 & \textbf{89.3} & \textbf{89.3} & 89.2 & \textbf{65.2} & 65.0 & 64.1 \\
VQA-X & 74.7 & \textbf{75.6} & 75.1 &\textbf{90.9} & 90.6 & 90.0 &\textbf{75.8} & 72.3 & 69.3 \\ 
\arrayrulecolor{black}\bottomrule
\end{tabular}
}
\caption{A comparison of \vatfive-CLIP model sizes (\sect{subsec:vision_lms}). $\dagger$ For VCR answers, we report both proxy accuracy (1st row) and  plausibility (2nd row; \sect{subsec:eval_metrics}).}
\label{tab:results_va_t5_scale}
\end{table*}

\paragraph{Results} 
We observe that CLIP features give the best accuracy scores  for all three datasets (Table \ref{tab:results_va_t5_features_accuracy}). %
This result demonstrates the benefit of visual adaptation: 
advances in image representations, such as CLIP, 
can be effortlessly used, unlike with joint models for which we need to re-train the model jointly from scratch with these new representations.   

In \sect{subsec:vision_lms}, we hypothesize that captioning could be a straightforward way to bridge two modalities to  get the most out of a PLM that is already
well-positioned to solve the end-task in one modality (text). %
However, with the exception of VQA-X, captions give the worst accuracy scores relative to object and CLIP features. 

We turn to evaluation of plausibility of generated explanations which paints a different picture (Table \ref{tab:results_va_t5_features_plausibility}). %
We observe that CLIP and object features perform similarly for VCR and VQA-X---object features are even slightly better for VQA-X. %
In other words, advances from CLIP features diminish for the more complex task of explanation generation. %
Captions work best for generating E-SNLI-VE explanations with VA-T5-Base, but not for the other two datasets. %
However, E-SNLI-VE is an outlier in another way.  %
It is the only task for which having no image features is better for explanation generation than having CLIP features,  and just slightly worse (0.3 points) than having object features. %
Notably, CLIP/object features require combining vectors from different models while captions are represented with the same pretrained word embeddings as the rest of the input. %
We thus explore whether the way that layer normalization is applied to the concatenated vectors is crucial, but we find that it is not (see Appendix). %
We leave further analysis for why visual adaptation gives only minor improvements for generation of E-SNLI-VE explanations with VA-T5 relative to other datasets to future work. 

BERTscore results (Table \ref{tab:results_va_t5_features_bertscore}) are mixed. %
According to them, CLIP features are the best for VCR and the worst for VQA-X. %
Moreover, the differences between BERTscore values obtained with different features are very small (0.0--0.4) which makes these results hard to interpret. 


\begin{table*}[t]
\small
    \centering
    \begin{subtable}{\textwidth}
    \resizebox{\textwidth}{!}{  
    \begin{tabular}{l|cccc|ccc|ccc}
    \toprule
          \multicolumn{1}{c}{} & \multicolumn{4}{c}{\textbf{\vcr}} & \multicolumn{3}{c}{\textbf{\esnlive}}       & \multicolumn{3}{c}{\textbf{\vqax}}           \\
         \multicolumn{1}{c}{} & \multicolumn{2}{c}{\emph{\small Answer}} & \multicolumn{2}{c}{\emph{\small  Explanation}} & \emph{\small  Answer} &  \multicolumn{2}{c}{\emph{\small  Explanation}} & \emph{\small  Answer} &  \multicolumn{2}{c}{\emph{\small  Explanation}} \\
          \cmidrule(lr){2-3}  \cmidrule(lr){4-5}  \cmidrule(lr){6-6} \cmidrule(lr){7-8} \cmidrule(lr){9-9} \cmidrule(lr){10-11}
          \multicolumn{1}{l}{\textbf{Model}} & \textbf{Acc.} & \textbf{Plaus.} & \textbf{BERTsc.} & \multicolumn{1}{c}{\textbf{Plaus.}} & \multicolumn{1}{c}{\textbf{Acc.}} & \textbf{BERTsc.}. & \multicolumn{1}{c}{\textbf{Plaus.}} & \multicolumn{1}{c}{\textbf{Acc.}} & \textbf{BERTsc.} & \textbf{Plaus.} \\
    \midrule
    
    VLP & 55.5 & \textbf{50.1} & 85.7 & \textbf{34.0} & 75.4 & 87.7 & 63.8 & 79.4 & 89.6 & 73.5 \\
    
    \arrayrulecolor{black!20}\midrule
    VA-T5-Base   & 58.1 & 41.4 & 86.0 & 21.0 & 74.7 & \textbf{89.3} & 65.2 & 74.7 & 90.9 & 75.8 \\
    VA-T5-Large  & \textbf{59.1} & 45.8 & 86.0 & 24.4 & 74.4 & 89.3 & 65.0 & 75.6 & 90.6 & 72.3 \\
    VA-T5-3B     & \textbf{59.1} & 42.2 & 85.8 & 25.8 & 68.0   & 89.2 & 64.1 & 75.1 & 90.0 & 69.3 \\
        \midrule
    VL-BART  & 57.8 & 47.7 & \textbf{86.6} & 29.5 & 75.6 & \textbf{89.3} & \textbf{71.5} & \textbf{86.3} & \textbf{91.2} & \textbf{75.9} \\
    VL-T5    & 58.4 & 44.6 & 85.5 & 28.7 & \textbf{76.3} & 89.1 & 69.0 & 84.9 & 91.0   & 72.2 \\
    \arrayrulecolor{black}\bottomrule
    \end{tabular}
    }
    \caption{High-resource data setting.}
    \label{tab:main_comparison_high_resource}
    \end{subtable}\par%
    \vspace*{0.5em}
    \begin{subtable}{\textwidth}
    \resizebox{\textwidth}{!}{
    \begin{tabular}{l|cccc|ccc|ccc}
    \toprule
          \multicolumn{1}{c}{} & \multicolumn{4}{c}{\textbf{\vcr}} & \multicolumn{3}{c}{\textbf{\esnlive}}       & \multicolumn{3}{c}{\textbf{\vqax}}           \\
         \multicolumn{1}{c}{} & \multicolumn{2}{c}{\emph{\small Answer}} & \multicolumn{2}{c}{\emph{\small  Explanation}} & \emph{\small  Answer} &  \multicolumn{2}{c}{\emph{\small  Explanation}} & \emph{\small  Answer} &  \multicolumn{2}{c}{\emph{\small  Explanation}} \\
          \cmidrule(lr){2-3}  \cmidrule(lr){4-5}  \cmidrule(lr){6-6} \cmidrule(lr){7-8} \cmidrule(lr){9-9} \cmidrule(lr){10-11}
          \multicolumn{1}{l}{\textbf{Model}} & \textbf{Acc.} & \textbf{Plaus.} & \textbf{BERTsc.} & \multicolumn{1}{c}{\textbf{Plaus.}} & \multicolumn{1}{c}{\textbf{Acc.}} & \textbf{BERTsc.}. & \multicolumn{1}{c}{\textbf{Plaus.}} & \multicolumn{1}{c}{\textbf{Acc.}} & \textbf{BERTsc.} & \textbf{Plaus.} \\
    \midrule
   VLP  & 54.7 & 21.1 & 85.9 & \textbf{25.1} & 73.5 & 87.6 &  63.6 & 71.8 & 89.4 & 72.9             \\
    \arrayrulecolor{black!20}\midrule
   VA-T5-Base  & 57.2 & 38.6 & 85.7 & 23.1 & 66.5 & 89.1 & 64.5 & 66.5 & 90.8 & \textbf{75.8}             \\
    VA-T5-Large & 57.3 & 37.3 & 85.8 & 21.1 & 68.3 & 89.2 & 62.0 & 67.3 & 90.4 & 73.6             \\
VA-T5-3B    & 57.0 & 32.5 & 85.3 & 19.0 & 68.7 & 89.2 & 63.6 & 53.7 & 90.6 & 69.8   \\
    \midrule
    VL-BART  & 57.4 & \textbf{42.8} & \textbf{86.4} & 22.7 & \textbf{74.4} & \textbf{89.3} & 66.1 & \textbf{85.6} & \textbf{90.9} & 72.9             \\
    VL-T5  & \textbf{58.5} & 41.5 & 85.2 & 22.5 & 74.2 & \textbf{89.3} & \textbf{66.6} & 83.5 & 90.5 & 71.3             \\
    \arrayrulecolor{black}\bottomrule
    \end{tabular}
    }
    \caption{Low-resource data setting.}
    \label{tab:main_comparison_low_resource}
    \end{subtable}
    \caption{Comparison of a joint VL mode (VLP), visually adapted pretrained LM (VA-T5), and combined models (VL-BART, VL-T5) on three datasets: \vcr, \esnlive, and \vqax. We report (proxy) \emph{answer} accuracy and plausibility (for VCR), and \emph{explanation} BERTscore and plausibility. See \sect{sec:models_background} for more information on models and \sect{sec:exp_setup} for tasks, datasets, and evaluation metrics.}
    \label{tab:main_comparison_all_data}
\end{table*}

\subsection{VA-T5: Analysis of Model Size}
\label{sec:va_t5_model_size}

Another advantage of visually adapting PLMs is that we can use larger model sizes since PLMs are typically more frequently scaled relative to joint models. %
The benefits of scaling the model and pretraining data size are outlined in Table \ref{tab:benefits_downsides}. %
We explore three model sizes for \vatfive (\sect{subsec:vision_lms}): Base (220M), Large (770M), and 3B. %
We use CLIP features to visually adapt T5 for these experiments since they give more accurate VA-T5 models, while generating explanations that are similarly plausible to those generated by T5 adapted with object features (see \sect{subsec:va_t5_features}). %
We use the full training sets to finetune VA-T5 models in this section.

\paragraph{Results} 

Scaling the model size from 220M (Base) to 770M (Large) parameters gives more accurate models for VCR and VQA-X, but further scaling to 3B parameters degrades performance. %
This is in contrast to self-rationalization of text-only inputs where performance monotonically increases with the T5's model size \cite{marasovic2022feb}.
E-SNLI-VE is an exception with no clear pattern between the model size and accuracy. %
Moreover, explanation plausibility decreases with the model size (Base > Large > 3B) for E-SNLI-VE and VQA-X. %
This is also in contrast to observations in text-only self-rationalization. 
The exact opposite is true for the plausibility of generated VCR explanations which increases with the model size (3B > Large > Base). %
Notably, in Table \ref{tab:benefits_downsides}, we report that the larger model and data size are correlated with capturing more world and commonsense knowledge, and generating VCR explanations requires more inferring about information that is unstated in the input relative to generating E-SNLI-VE or VQA-X explanations. %
This might explain the difference between why scaling is beneficial for VCR and not for other datasets.

Unlike accuracy and plausibility for which model scaling is helpful at least to some extent, BERTscore values decrease monotonically when scaling the model size. %
Despite reservations about this evaluation metric given its weak correlation with human judgements of explanation plausibility, BERTscore values are increasing monotonically for self-rationalization with textual inputs as expected \cite{marasovic2022feb}. %
Thus, we see this result as another evidence of the difficulty of visually adapting larger models rather than the limitations of BERTscore as an evaluation metric. 

A better understanding of what is the bottleneck for visually adapting larger PLMs is needed. %
We might need other ways to visually adapt besides the simple input changes that have been done so far.


\subsection{Joint Models vs.\ Visually Adapted PLMs}
\label{sec:main_comparison}
We turn to our main comparison between a joint model (VLP), a visually adapted PLM (VA-T5-CLIP), and  combined models (VL-BART, VL-T5). %
Given that joint models might be advantageous when finetuning data is limited, we compare the models when  finetuned with: (i) the entire training sets (high-resource data setting), and (ii) 30\% of the training data (low-resource data setting).

\paragraph{Results} 

In a high-resource setting (Table \ref{tab:main_comparison_high_resource}), the best \emph{answer} accuracy/plausibility is achieved by a different model for each dataset. %
To illustrate, VA-T5 (a visually adapted PLM) obtains the best VCR answer proxy accuracy, VLP (a joint model) VCR answer plausibility, VL-T5 (a combined model) E-SNLI-VE accuracy, and VL-BART (another combined model) VQA-X accuracy. %

Explanation plausibility results are slightly more consistent (Table \ref{tab:main_comparison_high_resource}). %
Namely, VL-BART generates most plausible explanations for E-SNLI-VE and VQA-X, and is only behind VLP for VCR. %
The best explanation BERTscore is also achieved with VL-BART for all tasks. %
However, the relative order of model types (joint, visually adapted, combined) across tasks is still mixed. %
Specifically, for VCR: joint > combined (both) > adapted (all); for E-SNLI-VE: combined (both) > adapted (all) > joint; for VQA-X: combined > adapted > joint > adapted > combined > adapted. %

It is not necessarily concerning that the results are mixed given the unique benefits and downsides of these models (see Table \ref{tab:benefits_downsides}) that could be relevant for one task and not another. %
However, observed cross-task differences in results are not intuitive. %
For example, visually adapting T5 and combined models give worse VCR explanation plausibility compared to plausibility obtained with the joint model, VLP. %
Since generating VCR explanations require the most commonsense and word knowledge relative to the other two tasks, it is reasonable to expect that this is a scenario where long pretraining using a more complex text will be beneficial, but it turns out it is not. 

We now turn to results in low-resource data setting (see Table \ref{tab:main_comparison_low_resource}). %
We hypothesized that joint models might work better when finetuning data is limited since there might not be enough images to appropriately visually adapt PLMs and they might still behave like (unimodal) language models. %
However, VA-T5 is not always underperforming relative to VLP, VL-BART, and VL-T5 in the low-resource setting. %
Specifically, VA-T5 is slightly better for explanation generation compared to VL-BART and VL-T5 for VCR, comparable to VLP for E-SNLI-VE, and better than all of the three models for VQA-X. %
It is also better than VLP for generating VCR answers. %
These results show that the size of finetuning data is not as detrimental for visual adaptation relative to joint models as we speculated.  

As in the high-resource setting, we observe that no model (type) works universally the best. %
Model ordering according to their performance sometimes stay consistent compare to the high-resource setting, e.g., for E-SNLI-VE accuracy, and sometimes changes notably, e.g., VLP gives the best VCR answer and explanation plausibility in high-resource setting, but the worst when data is limited. %
These results highlight the necessity to compare models, not only using a variety of tasks/datasets, but also models finetuned with different amounts of data.

We see that the differences between model performance in high-resource are smaller than in low-resource, where in some cases the gap is huge. %
For example, VL-BART achieves VCR answer plausibility of 42.8 in low-resource, while VLP results in only 21.1. %
Another example is VQA-X answer accuracy for which VL-BART achieves 85.6 and VLP 71.8. %
Unlike VCR, 
this can be explained by the fact that VQA is one of the tasks used to pretrain VL-BART (and VL-T5). 
Such huge differences between models are not observed for explanation generation, so even though a model is much more accurate, the plausibility of explanations for its correct answers are not that much more plausible compared to explanations of other models.

\section{Conclusions} 
We extensively analyze different multimodal models that have unique benefits and downsides for text generation conditioned on images and text beyond image captioning. %
We focus on self-rationalization (jointly generating labels/answers and free-text explanations), and show that there is no single approach that works best across instances of this complex domain. %
A key question moving forward is how best to leverage unimodal advances. 

In the meantime, our findings can be used as intermediate guidelines for which model to choose: 
\begin{compactitem}
\item Unlike for most text-only tasks, larger visually adapted language models, do not give better results. Our results suggest starting with T5-Large (770M parameters).
\item Although not always the best, CLIP features are a reasonable choice for visual adaptation. 
\item Do \emph{not} eliminate visual adaptation if your multimodal dataset is small.
\item If your multimodal data is \emph{not} limited, VL-BART is a reasonable baseline for multimodal self-rationalization. Otherwise, multiple models should be compared. 
\end{compactitem}


\section{Limitations}

While we examine multiple methods that were available to us while we were conducting this research, it is inevitable that new multimodal models will be released, leaving the question of whether those models are already superior for multimodal self-rationalizing open. %
For example, the recently proposed model OFA \cite{DBLP:conf/icml/WangYMLBLMZZY22} could be particularly suitable. %
It is available in large sizes (33M, 93M, 182M, 472M, 930M), trained with a filtered version of the large-scale text corpus PILE \cite[140GB;][]{DBLP:journals/corr/abs-2101-00027} as well as with a variety of multimodal datasets and objectives including image captioning and grounded captioning. 

Besides that, all of the models we examined have been trained on data in the English language and using clean, high quality images from similar data sources (MS COCO, Flickr). %
Inherent biases stemming from this source of data would need to be studied in future work towards scaling this work to multiple languages and other image sources (for e.g. noisy, dense context, adversarial images). %
Our main measure of explanation quality is plausibility that does not answer whether these plausible explanations are useful in real-world applications of VQA and NLI to actual stakeholders. 
Another limitation are our computational resources. With access to even more compute, we would be able to examine at a larger scale such as T5 11B or more.


\section*{Acknowledgments} 
The authors thank  members of the AllenNLP team at AI2 and anonymous reviewers for
their helpful feedback, and Maxime Kayser for releasing the crowdsourcing templates. 

\bibliography{anthology, custom}

\begin{thebibliography}{60}
\expandafter\ifx\csname natexlab\endcsname\relax\def\natexlab#1{#1}\fi

\bibitem[{Antol et~al.(2015)Antol, Agrawal, Lu, Mitchell, Batra, Zitnick, and
  Parikh}]{antol2015vqa}
Stanislaw Antol, Aishwarya Agrawal, Jiasen Lu, Margaret Mitchell, Dhruv Batra,
  C~Lawrence Zitnick, and Devi Parikh. 2015.
\newblock \href {https://arxiv.org/abs/1505.00468} {Vqa: Visual question
  answering}.
\newblock In \emph{Proceedings of the IEEE/CVF International Conference on
  Computer Vision (ICCV)}.

\bibitem[{Bowman et~al.(2015)Bowman, Angeli, Potts, and
  Manning}]{bowman-etal-2015-large}
Samuel~R. Bowman, Gabor Angeli, Christopher Potts, and Christopher~D. Manning.
  2015.
\newblock \href {https://doi.org/10.18653/v1/D15-1075} {A large annotated
  corpus for learning natural language inference}.
\newblock In \emph{Proceedings of the 2015 Conference on Empirical Methods in
  Natural Language Processing}, pages 632--642, Lisbon, Portugal. Association
  for Computational Linguistics.

\bibitem[{Brown et~al.(2020)Brown, Mann, Ryder, Subbiah, Kaplan, Dhariwal,
  Neelakantan, Shyam, Sastry, Askell, Agarwal, Herbert-Voss, Krueger, Henighan,
  Child, Ramesh, Ziegler, Wu, Winter, Hesse, Chen, Sigler, Litwin, Gray, Chess,
  Clark, Berner, McCandlish, Radford, Sutskever, and Amodei}]{brown2020gpt3}
Tom Brown, Benjamin Mann, Nick Ryder, Melanie Subbiah, Jared~D Kaplan, Prafulla
  Dhariwal, Arvind Neelakantan, Pranav Shyam, Girish Sastry, Amanda Askell,
  Sandhini Agarwal, Ariel Herbert-Voss, Gretchen Krueger, Tom Henighan, Rewon
  Child, Aditya Ramesh, Daniel Ziegler, Jeffrey Wu, Clemens Winter, Chris
  Hesse, Mark Chen, Eric Sigler, Mateusz Litwin, Scott Gray, Benjamin Chess,
  Jack Clark, Christopher Berner, Sam McCandlish, Alec Radford, Ilya Sutskever,
  and Dario Amodei. 2020.
\newblock \href
  {https://proceedings.neurips.cc/paper/2020/file/1457c0d6bfcb4967418bfb8ac142f64a-Paper.pdf}
  {Language models are few-shot learners}.
\newblock In \emph{Advances in Neural Information Processing Systems},
  volume~33, pages 1877--1901. Curran Associates, Inc.

\bibitem[{Camburu et~al.(2018)Camburu, Rockt{\"a}schel, Lukasiewicz, and
  Blunsom}]{camburu2018snli}
Oana-Maria Camburu, Tim Rockt{\"a}schel, Thomas Lukasiewicz, and Phil Blunsom.
  2018.
\newblock \href
  {https://papers.nips.cc/paper/8163-e-snli-natural-language-inference-with-natural-language-explanations/}
  {{e-SNLI: N}atural language inference with natural language explanations}.
\newblock In \emph{Advances in Neural Information Processing Systems
  (NeurIPS)}.

\bibitem[{Chang et~al.(2022)Chang, Narang, Suzuki, Cao, Gao, and
  Bisk}]{Chang2022}
Yingshan Chang, Mridu Narang, Hisami Suzuki, Guihong Cao, Jianfeng Gao, and
  Yonatan Bisk. 2022.
\newblock \href {https://arxiv.org/abs/2109.00590} {{WebQA: Multihop and
  Multimodal QA}}.
\newblock In \emph{Conference on Computer Vision and Pattern Recognition}.

\bibitem[{Chen et~al.(2019)Chen, Li, Yu, Kholy, Ahmed, Gan, Cheng, and
  Liu}]{Chen2019UNITERLU}
Yen-Chun Chen, Linjie Li, Licheng Yu, Ahmed~El Kholy, Faisal Ahmed, Zhe Gan,
  Yu~Cheng, and Jingjing Liu. 2019.
\newblock \href {https://arxiv.org/abs/1909.11740} {{UNITER: Learning UNiversal
  Image-TExt Representations}}.
\newblock In \emph{Proceedings of the European Conference on Computer Vision
  (ECCV)}.

\bibitem[{Cho et~al.(2021)Cho, Lei, Tan, and Bansal}]{pmlr-v139-cho21a}
Jaemin Cho, Jie Lei, Hao Tan, and Mohit Bansal. 2021.
\newblock \href {https://proceedings.mlr.press/v139/cho21a.html} {Unifying
  vision-and-language tasks via text generation}.
\newblock In \emph{Proceedings of the 38th International Conference on Machine
  Learning}. PMLR.

\bibitem[{Clinciu et~al.(2021)Clinciu, Eshghi, and
  Hastie}]{clinciu-etal-2021-study}
Miruna-Adriana Clinciu, Arash Eshghi, and Helen Hastie. 2021.
\newblock \href {https://doi.org/10.18653/v1/2021.eacl-main.202} {A study of
  automatic metrics for the evaluation of natural language explanations}.
\newblock In \emph{Proceedings of the 16th Conference of the European Chapter
  of the Association for Computational Linguistics: Main Volume}, pages
  2376--2387, Online. Association for Computational Linguistics.

\bibitem[{Davison et~al.(2019)Davison, Feldman, and
  Rush}]{davison-etal-2019-commonsense}
Joe Davison, Joshua Feldman, and Alexander Rush. 2019.
\newblock \href {https://doi.org/10.18653/v1/D19-1109} {Commonsense knowledge
  mining from pretrained models}.
\newblock In \emph{Proceedings of the 2019 Conference on Empirical Methods in
  Natural Language Processing and the 9th International Joint Conference on
  Natural Language Processing (EMNLP-IJCNLP)}, pages 1173--1178, Hong Kong,
  China. Association for Computational Linguistics.

\bibitem[{Devlin et~al.(2019)Devlin, Chang, Lee, and
  Toutanova}]{devlin-etal-2019-bert}
Jacob Devlin, Ming-Wei Chang, Kenton Lee, and Kristina Toutanova. 2019.
\newblock \href {https://doi.org/10.18653/v1/N19-1423} {{BERT}: Pre-training of
  deep bidirectional transformers for language understanding}.
\newblock In \emph{Proceedings of the 2019 Conference of the North {A}merican
  Chapter of the Association for Computational Linguistics: Human Language
  Technologies, Volume 1 (Long and Short Papers)}, pages 4171--4186,
  Minneapolis, Minnesota. Association for Computational Linguistics.

\bibitem[{Dodge et~al.(2021)Dodge, Sap, Marasovi{\'c}, Agnew, Ilharco,
  Groeneveld, Mitchell, and Gardner}]{dodge-etal-2021-documenting}
Jesse Dodge, Maarten Sap, Ana Marasovi{\'c}, William Agnew, Gabriel Ilharco,
  Dirk Groeneveld, Margaret Mitchell, and Matt Gardner. 2021.
\newblock \href {https://doi.org/10.18653/v1/2021.emnlp-main.98} {Documenting
  large webtext corpora: A case study on the colossal clean crawled corpus}.
\newblock In \emph{Proceedings of the 2021 Conference on Empirical Methods in
  Natural Language Processing}, pages 1286--1305, Online and Punta Cana,
  Dominican Republic. Association for Computational Linguistics.

\bibitem[{Dong et~al.(2019)Dong, Yang, Wang, Wei, Liu, Wang, Gao, Zhou, and
  Hon}]{DBLP:conf/nips/00040WWLWGZH19}
Li~Dong, Nan Yang, Wenhui Wang, Furu Wei, Xiaodong Liu, Yu~Wang, Jianfeng Gao,
  Ming Zhou, and Hsiao{-}Wuen Hon. 2019.
\newblock \href
  {https://proceedings.neurips.cc/paper/2019/hash/c20bb2d9a50d5ac1f713f8b34d9aac5a-Abstract.html}
  {Unified language model pre-training for natural language understanding and
  generation}.
\newblock In \emph{Advances in Neural Information Processing Systems 32: Annual
  Conference on Neural Information Processing Systems 2019, NeurIPS 2019,
  December 8-14, 2019, Vancouver, BC, Canada}, pages 13042--13054.

\bibitem[{Dosovitskiy et~al.(2021)Dosovitskiy, Beyer, Kolesnikov, Weissenborn,
  Zhai, Unterthiner, Dehghani, Minderer, Heigold, Gelly, Uszkoreit, and
  Houlsby}]{DBLP:conf/iclr/DosovitskiyB0WZ21}
Alexey Dosovitskiy, Lucas Beyer, Alexander Kolesnikov, Dirk Weissenborn,
  Xiaohua Zhai, Thomas Unterthiner, Mostafa Dehghani, Matthias Minderer, Georg
  Heigold, Sylvain Gelly, Jakob Uszkoreit, and Neil Houlsby. 2021.
\newblock \href {https://openreview.net/forum?id=YicbFdNTTy} {An image is worth
  16x16 words: Transformers for image recognition at scale}.
\newblock In \emph{9th International Conference on Learning Representations,
  {ICLR} 2021, Virtual Event, Austria, May 3-7, 2021}. OpenReview.net.

\bibitem[{Dua et~al.(2021)Dua, Kancheti, and Balasubramanian}]{Dua2021BeyondVG}
Radhika Dua, Sai~Srinivas Kancheti, and Vineeth~N. Balasubramanian. 2021.
\newblock \href {https://arxiv.org/abs/2010.12852} {Beyond vqa: Generating
  multi-word answers and rationales to visual questions}.
\newblock In \emph{Proceedings of IEEE/CVF Conference on Computer Vision and
  Pattern Recognition Workshops}.

\bibitem[{Eichenberg et~al.(2021)Eichenberg, Black, Weinbach, Parcalabescu, and
  Frank}]{Eichenberg2021MAGMAM}
Constantin Eichenberg, Sid Black, Samuel Weinbach, Letitia Parcalabescu, and
  Anette Frank. 2021.
\newblock \href {https://arxiv.org/abs/2112.05253} {Magma - multimodal
  augmentation of generative models through adapter-based finetuning}.
\newblock {arXiv:abs/2112.05253}.

\bibitem[{Gao et~al.(2021)Gao, Biderman, Black, Golding, Hoppe, Foster, Phang,
  He, Thite, Nabeshima, Presser, and Leahy}]{DBLP:journals/corr/abs-2101-00027}
Leo Gao, Stella Biderman, Sid Black, Laurence Golding, Travis Hoppe, Charles
  Foster, Jason Phang, Horace He, Anish Thite, Noa Nabeshima, Shawn Presser,
  and Connor Leahy. 2021.
\newblock \href {http://arxiv.org/abs/2101.00027} {The pile: An 800gb dataset
  of diverse text for language modeling}.
\newblock \emph{CoRR}, abs/2101.00027.

\bibitem[{Gokaslan and Cohen(2019)}]{Gokaslan2019OpenWeb}
Aaron Gokaslan and Vanya Cohen. 2019.
\newblock \href {http://Skylion007.github.io/OpenWebTextCorpus} {Openwebtext
  corpus}.

\bibitem[{Goyal et~al.(2017)Goyal, Khot, Summers-Stay, Batra, and
  Parikh}]{goyal2017making}
Yash Goyal, Tejas Khot, Douglas Summers-Stay, Dhruv Batra, and Devi Parikh.
  2017.
\newblock \href {https://arxiv.org/abs/1612.00837} {Making the {V} in {VQA}
  matter: {E}levating the role of image understanding in visual question
  answering}.
\newblock In \emph{Proceedings of the IEEE Conference on Computer Vision and
  Pattern Recognition}.

\bibitem[{Gui et~al.(2022{\natexlab{a}})Gui, Huang, Hauptmann, Bisk, and
  Gao}]{Gui2022}
Liangke Gui, Qiuyuan Huang, Alex Hauptmann, Yonatan Bisk, and Jianfeng Gao.
  2022{\natexlab{a}}.
\newblock \href {https://arxiv.org/abs/2205.09256} {{Training Vision-Language
  Transformers from Captions Alone}}.
\newblock \emph{ArXiv}.

\bibitem[{Gui et~al.(2022{\natexlab{b}})Gui, Wang, Huang, Hauptmann, Bisk, and
  Gao}]{Gui2022KAT}
Liangke Gui, Borui Wang, Qiuyuan Huang, Alex Hauptmann, Yonatan Bisk, and
  Jianfeng Gao. 2022{\natexlab{b}}.
\newblock \href {https://arxiv.org/abs/2112.08614} {{KAT: A Knowledge Augmented
  Transformer for Vision-and-Language}}.
\newblock In \emph{Annual Conference of the North American Chapter of the
  Association for Computational Linguistics}.

\bibitem[{Gupta et~al.(2022)Gupta, Kamath, Kembhavi, and
  Hoiem}]{gupta2021towards}
Tanmay Gupta, Amita Kamath, Aniruddha Kembhavi, and Derek Hoiem. 2022.
\newblock \href {https://arxiv.org/abs/2104.00743} {Towards general purpose
  vision systems}.
\newblock In \emph{The IEEE / CVF Computer Vision and Pattern Recognition
  Conference (CVPR)}.

\bibitem[{Hase et~al.(2020)Hase, Zhang, Xie, and
  Bansal}]{hase-etal-2020-leakage}
Peter Hase, Shiyue Zhang, Harry Xie, and Mohit Bansal. 2020.
\newblock \href {https://doi.org/10.18653/v1/2020.findings-emnlp.390}
  {Leakage-adjusted simulatability: Can models generate non-trivial
  explanations of their behavior in natural language?}
\newblock In \emph{Findings of the Association for Computational Linguistics:
  EMNLP 2020}, pages 4351--4367, Online. Association for Computational
  Linguistics.

\bibitem[{Hudson and Manning(2019)}]{hudson2019gqa}
Drew~A Hudson and Christopher~D Manning. 2019.
\newblock \href {https://arxiv.org/abs/1902.09506} {Gqa: A new dataset for
  real-world visual reasoning and compositional question answering}.
\newblock In \emph{Proceedings of the IEEE/CVF Conference on Computer Vision
  and Pattern Recognition}.

\bibitem[{Kayser et~al.(2021)Kayser, Camburu, Salewski, Emde, Do, Akata, and
  Lukasiewicz}]{Kayser2021eViLAD}
Maxime Kayser, Oana-Maria Camburu, Leonard Salewski, Cornelius Emde, Virginie
  Do, Zeynep Akata, and Thomas Lukasiewicz. 2021.
\newblock \href
  {https://openaccess.thecvf.com/content/ICCV2021/papers/Kayser_E-ViL_A_Dataset_and_Benchmark_for_Natural_Language_Explanations_in_ICCV_2021_paper.pdf}
  {e-vil: A dataset and benchmark for natural language explanations in
  vision-language tasks}.
\newblock In \emph{Proceedings of the IEEE/CVF International Conference on
  Computer Vision (ICCV)}.

\bibitem[{Krishna et~al.(2017)Krishna, Zhu, Groth, Johnson, Hata, Kravitz,
  Chen, Kalantidis, Li, Shamma, Bernstein, and Fei{-}Fei}]{krishna2017visual}
Ranjay Krishna, Yuke Zhu, Oliver Groth, Justin Johnson, Kenji Hata, Joshua
  Kravitz, Stephanie Chen, Yannis Kalantidis, Li{-}Jia Li, David~A. Shamma,
  Michael~S. Bernstein, and Li~Fei{-}Fei. 2017.
\newblock \href {https://doi.org/10.1007/s11263-016-0981-7} {Visual genome:
  Connecting language and vision using crowdsourced dense image annotations}.
\newblock \emph{Int. J. Comput. Vis.}, 123(1):32--73.

\bibitem[{Lewis et~al.(2020)Lewis, Liu, Goyal, Ghazvininejad, Mohamed, Levy,
  Stoyanov, and Zettlemoyer}]{lewis-etal-2020-bart}
Mike Lewis, Yinhan Liu, Naman Goyal, Marjan Ghazvininejad, Abdelrahman Mohamed,
  Omer Levy, Veselin Stoyanov, and Luke Zettlemoyer. 2020.
\newblock \href {https://doi.org/10.18653/v1/2020.acl-main.703} {{BART}:
  Denoising sequence-to-sequence pre-training for natural language generation,
  translation, and comprehension}.
\newblock In \emph{Proceedings of the 58th Annual Meeting of the Association
  for Computational Linguistics}, pages 7871--7880, Online. Association for
  Computational Linguistics.

\bibitem[{Li et~al.(2020{\natexlab{a}})Li, Duan, Fang, Jiang, and
  Zhou}]{Li2019UnicoderVLAU}
Gen Li, Nan Duan, Yuejian Fang, Daxin Jiang, and Ming Zhou. 2020{\natexlab{a}}.
\newblock \href {https://arxiv.org/abs/1908.06066} {{Unicoder-VL: A Universal
  Encoder for Vision and Language by Cross-modal Pre-training}}.
\newblock In \emph{AAAI}.

\bibitem[{Li et~al.(2020{\natexlab{b}})Li, Yin, Li, Zhang, Hu, Zhang, Wang, Hu,
  Dong, Wei, Choi, and Gao}]{li2020oscar}
Xiujun Li, Xi~Yin, Chunyuan Li, Pengchuan Zhang, Xiaowei Hu, Lei Zhang, Lijuan
  Wang, Houdong Hu, Li~Dong, Furu Wei, Yejin Choi, and Jianfeng Gao.
  2020{\natexlab{b}}.
\newblock \href
  {https://www.ecva.net/papers/eccv_2020/papers_ECCV/papers/123750120.pdf}
  {Oscar: Object-semantics aligned pre-training for vision-language tasks}.
\newblock In \emph{Proceedings of the European Conference on Computer Vision
  (ECCV)}.

\bibitem[{Lin et~al.(2014)Lin, Maire, Belongie, Hays, Perona, Ramanan,
  Doll{\'a}r, and Zitnick}]{lin2014microsoft}
Tsung-Yi Lin, Michael Maire, Serge Belongie, James Hays, Pietro Perona, Deva
  Ramanan, Piotr Doll{\'a}r, and C~Lawrence Zitnick. 2014.
\newblock \href {https://arxiv.org/abs/1405.0312} {Microsoft coco: Common
  objects in context}.
\newblock In \emph{Proceedings of the European Conference on Computer Vision
  (ECCV)}.

\bibitem[{Lu et~al.(2019)Lu, Batra, Parikh, and Lee}]{Lu2019ViLBERTPT}
Jiasen Lu, Dhruv Batra, Devi Parikh, and Stefan Lee. 2019.
\newblock \href {https://arxiv.org/abs/1908.02265} {{ViLBERT: Pretraining
  Task-Agnostic Visiolinguistic Representations for Vision-and-Language
  Tasks}}.
\newblock In \emph{NeurIPS}.

\bibitem[{Marasovi{\'c} et~al.(2022)Marasovi{\'c}, Beltagy, Downey, and
  Peters}]{marasovic2022feb}
Ana Marasovi{\'c}, Iz~Beltagy, Doug Downey, and Matthew~E. Peters. 2022.
\newblock \href {https://arxiv.org/abs/2111.08284} {Few-shot
  self-rationalization with natural language prompts}.
\newblock In \emph{Findings of the Association for Computational Linguistics:
  NAACL 2022}.

\bibitem[{Marasovi{\'c} et~al.(2020)Marasovi{\'c}, Bhagavatula, Park, Le~Bras,
  Smith, and Choi}]{marasovic-etal-2020-natural}
Ana Marasovi{\'c}, Chandra Bhagavatula, Jae~sung Park, Ronan Le~Bras, Noah~A.
  Smith, and Yejin Choi. 2020.
\newblock \href {https://doi.org/10.18653/v1/2020.findings-emnlp.253} {Natural
  language rationales with full-stack visual reasoning: From pixels to semantic
  frames to commonsense graphs}.
\newblock In \emph{Findings of the Association for Computational Linguistics:
  EMNLP 2020}, pages 2810--2829, Online. Association for Computational
  Linguistics.

\bibitem[{Narang et~al.(2020)Narang, Raffel, Lee, Roberts, Fiedel, and
  Malkan}]{DBLP:journals/corr/abs-2004-14546}
Sharan Narang, Colin Raffel, Katherine Lee, Adam Roberts, Noah Fiedel, and
  Karishma Malkan. 2020.
\newblock \href {http://arxiv.org/abs/2004.14546} {Wt5?! training text-to-text
  models to explain their predictions}.
\newblock \emph{CoRR}, abs/2004.14546.

\bibitem[{Park et~al.(2018)Park, Hendricks, Akata, Rohrbach, Schiele, Darrell,
  and Rohrbach}]{Park2018MultimodalEJ}
Dong~Huk Park, Lisa~Anne Hendricks, Zeynep Akata, Anna Rohrbach, Bernt Schiele,
  Trevor Darrell, and Marcus Rohrbach. 2018.
\newblock \href
  {https://openaccess.thecvf.com/content_cvpr_2018/papers/Park_Multimodal_Explanations_Justifying_CVPR_2018_paper.pdf}
  {Multimodal explanations: Justifying decisions and pointing to the evidence}.
\newblock In \emph{Proceedings of the IEEE/CVF Conference on Computer Vision
  and Pattern Recognition (CVPR)}.

\bibitem[{Park et~al.(2020)Park, Bhagavatula, Mottaghi, Farhadi, and
  Choi}]{Park2020VisualCG}
Jae~Sung Park, Chandra Bhagavatula, Roozbeh Mottaghi, Ali Farhadi, and Yejin
  Choi. 2020.
\newblock \href {https://arxiv.org/abs/2004.10796} {{Visual Commonsense Graphs:
  Reasoning about the Dynamic Context of a Still Image}}.
\newblock In \emph{Proceedings of the European Conference on Computer Vision
  (ECCV)}.

\bibitem[{Petroni et~al.(2019)Petroni, Rockt{\"a}schel, Riedel, Lewis, Bakhtin,
  Wu, and Miller}]{petroni-etal-2019-language}
Fabio Petroni, Tim Rockt{\"a}schel, Sebastian Riedel, Patrick Lewis, Anton
  Bakhtin, Yuxiang Wu, and Alexander Miller. 2019.
\newblock \href {https://doi.org/10.18653/v1/D19-1250} {Language models as
  knowledge bases?}
\newblock In \emph{Proceedings of the 2019 Conference on Empirical Methods in
  Natural Language Processing and the 9th International Joint Conference on
  Natural Language Processing (EMNLP-IJCNLP)}, pages 2463--2473, Hong Kong,
  China. Association for Computational Linguistics.

\bibitem[{Radford et~al.(2021)Radford, Kim, Hallacy, Ramesh, Goh, Agarwal,
  Sastry, Askell, Mishkin, Clark, Krueger, and
  Sutskever}]{pmlr-v139-radford21a}
Alec Radford, Jong~Wook Kim, Chris Hallacy, Aditya Ramesh, Gabriel Goh,
  Sandhini Agarwal, Girish Sastry, Amanda Askell, Pamela Mishkin, Jack Clark,
  Gretchen Krueger, and Ilya Sutskever. 2021.
\newblock \href {https://proceedings.mlr.press/v139/radford21a.html} {Learning
  transferable visual models from natural language supervision}.
\newblock In \emph{Proceedings of the 38th International Conference on Machine
  Learning}, volume 139 of \emph{Proceedings of Machine Learning Research},
  pages 8748--8763. PMLR.

\bibitem[{Radford et~al.(2019)Radford, Wu, Child, Luan, Amodei, and
  Sutskever}]{Radford2019LanguageMA}
Alec Radford, Jeffrey Wu, Rewon Child, David Luan, Dario Amodei, and Ilya
  Sutskever. 2019.
\newblock \href
  {https://cdn.openai.com/better-language-models/language_models_are_unsupervised_multitask_learners.pdf}
  {{Language Models are Unsupervised Multitask Learners}}.

\bibitem[{Raffel et~al.(2020)Raffel, Shazeer, Roberts, Lee, Narang, Matena,
  Zhou, Li, and Liu}]{raffel2019exploring}
Colin Raffel, Noam Shazeer, Adam Roberts, Katherine Lee, Sharan Narang, Michael
  Matena, Yanqi Zhou, Wei Li, and Peter~J. Liu. 2020.
\newblock \href {http://jmlr.org/papers/v21/20-074.html} {Exploring the limits
  of transfer learning with a unified text-to-text transformer}.
\newblock \emph{Journal of Machine Learning Research}, 21(140):1--67.

\bibitem[{Ren et~al.(2015)Ren, He, Girshick, and Sun}]{Ren2015FasterRT}
Shaoqing Ren, Kaiming He, Ross~B. Girshick, and Jian Sun. 2015.
\newblock \href {https://arxiv.org/abs/1506.01497} {{Faster R-CNN: Towards
  Real-Time Object Detection with Region Proposal Networks}}.
\newblock \emph{IEEE Transactions on Pattern Analysis and Machine
  Intelligence}, 39:1137--1149.

\bibitem[{Rohrbach et~al.(2017)Rohrbach, Torabi, Rohrbach, Tandon, Pal,
  Larochelle, Courville, and Schiele}]{DBLP:journals/ijcv/RohrbachTRTPLCS17}
Anna Rohrbach, Atousa Torabi, Marcus Rohrbach, Niket Tandon, Christopher~Joseph
  Pal, Hugo Larochelle, Aaron~C. Courville, and Bernt Schiele. 2017.
\newblock \href {https://doi.org/10.1007/s11263-016-0987-1} {Movie
  description}.
\newblock \emph{Int. J. Comput. Vis.}, 123(1):94--120.

\bibitem[{Sharma et~al.(2018)Sharma, Ding, Goodman, and
  Soricut}]{sharma2018conceptual}
Piyush Sharma, Nan Ding, Sebastian Goodman, and Radu Soricut. 2018.
\newblock \href {https://doi.org/10.18653/v1/P18-1238} {Conceptual captions: A
  cleaned, hypernymed, image alt-text dataset for automatic image captioning}.
\newblock In \emph{Proceedings of the 56th Annual Meeting of the Association
  for Computational Linguistics (Volume 1: Long Papers)}, pages 2556--2565,
  Melbourne, Australia. Association for Computational Linguistics.

\bibitem[{Shen et~al.(2022)Shen, Li, Tan, Bansal, Rohrbach, Chang, Yao, and
  Keutzer}]{Shen2021HowMC}
Sheng Shen, Liunian~Harold Li, Hao Tan, Mohit Bansal, Anna Rohrbach, Kai-Wei
  Chang, Zhewei Yao, and Kurt Keutzer. 2022.
\newblock \href {https://arxiv.org/abs/2107.06383} {How much can clip benefit
  vision-and-language tasks?}
\newblock In \emph{ICLR}.

\bibitem[{Sollami and Jain(2021)}]{Sollami2021MultimodalCF}
Michael Sollami and Aashish Jain. 2021.
\newblock \href {https://arxiv.org/abs/2109.01229} {Multimodal conditionality
  for natural language generation}.
\newblock {arXiv:2109.01229}.

\bibitem[{Su et~al.(2020)Su, Zhu, Cao, Li, Lu, Wei, and Dai}]{Su2019VLBERTPO}
Weijie Su, Xizhou Zhu, Yue Cao, Bin Li, Lewei Lu, Furu Wei, and Jifeng Dai.
  2020.
\newblock \href {https://arxiv.org/abs/1908.08530} {{VL-BERT: Pre-training of
  Generic Visual-Linguistic Representations}}.
\newblock In \emph{ICLR}.

\bibitem[{Tan and Bansal(2019)}]{tan-bansal-2019-lxmert}
Hao Tan and Mohit Bansal. 2019.
\newblock \href {https://doi.org/10.18653/v1/D19-1514} {{LXMERT}: Learning
  cross-modality encoder representations from transformers}.
\newblock In \emph{Proceedings of the 2019 Conference on Empirical Methods in
  Natural Language Processing and the 9th International Joint Conference on
  Natural Language Processing (EMNLP-IJCNLP)}, pages 5100--5111, Hong Kong,
  China. Association for Computational Linguistics.

\bibitem[{Thomee et~al.(2016)Thomee, Shamma, Friedland, Elizalde, Ni, Poland,
  Borth, and Li}]{DBLP:journals/cacm/ThomeeSFENPBL16}
Bart Thomee, David~A. Shamma, Gerald Friedland, Benjamin Elizalde, Karl Ni,
  Douglas Poland, Damian Borth, and Li{-}Jia Li. 2016.
\newblock \href {https://doi.org/10.1145/2812802} {{YFCC100M:} the new data in
  multimedia research}.
\newblock \emph{Commun. {ACM}}, 59(2):64--73.

\bibitem[{Trinh and Le(2018)}]{DBLP:journals/corr/abs-1806-02847}
Trieu~H. Trinh and Quoc~V. Le. 2018.
\newblock \href {http://arxiv.org/abs/1806.02847} {A simple method for
  commonsense reasoning}.
\newblock \emph{CoRR}, abs/1806.02847.

\bibitem[{Vaswani et~al.(2017)Vaswani, Shazeer, Parmar, Uszkoreit, Jones,
  Gomez, Kaiser, and Polosukhin}]{vaswani2017attention}
Ashish Vaswani, Noam Shazeer, Niki Parmar, Jakob Uszkoreit, Llion Jones,
  Aidan~N Gomez, {\L}ukasz Kaiser, and Illia Polosukhin. 2017.
\newblock \href {https://arxiv.org/abs/1706.03762} {{Attention Is All You
  Need}}.
\newblock In \emph{NeurIPS}.

\bibitem[{Wang et~al.(2022{\natexlab{a}})Wang, Yang, Men, Lin, Bai, Li, Ma,
  Zhou, Zhou, and Yang}]{DBLP:conf/icml/WangYMLBLMZZY22}
Peng Wang, An~Yang, Rui Men, Junyang Lin, Shuai Bai, Zhikang Li, Jianxin Ma,
  Chang Zhou, Jingren Zhou, and Hongxia Yang. 2022{\natexlab{a}}.
\newblock \href {https://proceedings.mlr.press/v162/wang22al.html} {{OFA:}
  unifying architectures, tasks, and modalities through a simple
  sequence-to-sequence learning framework}.
\newblock In \emph{International Conference on Machine Learning, {ICML} 2022,
  17-23 July 2022, Baltimore, Maryland, {USA}}, volume 162 of \emph{Proceedings
  of Machine Learning Research}, pages 23318--23340. {PMLR}.

\bibitem[{Wang et~al.(2022{\natexlab{b}})Wang, Yu, Yu, Dai, Tsvetkov, and
  Cao}]{wang2022SimVLM}
Zirui Wang, Jiahui Yu, Adams~Wei Yu, Zihang Dai, Yulia Tsvetkov, and Yuan Cao.
  2022{\natexlab{b}}.
\newblock \href {https://arxiv.org/abs/2108.10904} {Simvlm: Simple visual
  language model pretraining with weak supervision}.
\newblock In \emph{ICLR}.

\bibitem[{Wiegreffe et~al.(2021)Wiegreffe, Marasovi{\'c}, and
  Smith}]{wiegreffe-etal-2021-measuring}
Sarah Wiegreffe, Ana Marasovi{\'c}, and Noah~A. Smith. 2021.
\newblock \href {https://doi.org/10.18653/v1/2021.emnlp-main.804} {{M}easuring
  association between labels and free-text rationales}.
\newblock In \emph{Proceedings of the 2021 Conference on Empirical Methods in
  Natural Language Processing}, pages 10266--10284, Online and Punta Cana,
  Dominican Republic. Association for Computational Linguistics.

\bibitem[{Wu et~al.(2019)Wu, Kirillov, Massa, Lo, and
  Girshick}]{wu2019detectron2}
Yuxin Wu, Alexander Kirillov, Francisco Massa, Wan-Yen Lo, and Ross Girshick.
  2019.
\newblock Detectron2.
\newblock \url{https://github.com/facebookresearch/detectron2}.

\bibitem[{Xie et~al.(2019)Xie, Lai, Doran, and Kadav}]{xie2019visual}
Ning Xie, Farley Lai, Derek Doran, and Asim Kadav. 2019.
\newblock \href {https://arxiv.org/pdf/1901.06706.pdf} {Visual entailment: A
  novel task for fine-grained image understanding}.
\newblock \emph{arXiv preprint arXiv:1901.06706}.

\bibitem[{Xie et~al.(2017)Xie, Girshick, Doll{\'{a}}r, Tu, and
  He}]{DBLP:conf/cvpr/XieGDTH17}
Saining Xie, Ross~B. Girshick, Piotr Doll{\'{a}}r, Zhuowen Tu, and Kaiming He.
  2017.
\newblock \href {https://doi.org/10.1109/CVPR.2017.634} {Aggregated residual
  transformations for deep neural networks}.
\newblock In \emph{2017 {IEEE} Conference on Computer Vision and Pattern
  Recognition, {CVPR} 2017, Honolulu, HI, USA, July 21-26, 2017}, pages
  5987--5995. {IEEE} Computer Society.

\bibitem[{Young et~al.(2014)Young, Lai, Hodosh, and
  Hockenmaier}]{young-etal-2014-image}
Peter Young, Alice Lai, Micah Hodosh, and Julia Hockenmaier. 2014.
\newblock \href {https://doi.org/10.1162/tacl_a_00166} {From image descriptions
  to visual denotations: New similarity metrics for semantic inference over
  event descriptions}.
\newblock \emph{Transactions of the Association for Computational Linguistics},
  2:67--78.

\bibitem[{Zellers et~al.(2019)Zellers, Bisk, Farhadi, and
  Choi}]{Zellers2019FromRT}
Rowan Zellers, Yonatan Bisk, Ali Farhadi, and Yejin Choi. 2019.
\newblock \href
  {https://openaccess.thecvf.com/content_CVPR_2019/papers/Zellers_From_Recognition_to_Cognition_Visual_Commonsense_Reasoning_CVPR_2019_paper.pdf}
  {From recognition to cognition: Visual commonsense reasoning}.
\newblock In \emph{Proceedings of the IEEE/CVF Conference on Computer Vision
  and Pattern Recognition (CVPR)}.

\bibitem[{Zhou et~al.(2020)Zhou, Palangi, Zhang, Hu, Corso, and
  Gao}]{zhou2020unified}
Luowei Zhou, Hamid Palangi, Lei Zhang, Houdong Hu, Jason Corso, and Jianfeng
  Gao. 2020.
\newblock \href {https://arxiv.org/abs/1909.11059} {Unified vision-language
  pre-training for image captioning and vqa}.
\newblock In \emph{Proceedings of the AAAI Conference on Artificial
  Intelligence}.

\bibitem[{Zhu et~al.(2016)Zhu, Groth, Bernstein, and Fei-Fei}]{zhu2016visual7w}
Yuke Zhu, Oliver Groth, Michael Bernstein, and Li~Fei-Fei. 2016.
\newblock \href {https://arxiv.org/abs/1511.03416} {Visual7w: Grounded question
  answering in images}.
\newblock In \emph{Proceedings of the IEEE Conference on Computer Vision and
  Pattern Recognition}.

\bibitem[{Zhu et~al.(2015)Zhu, Kiros, Zemel, Salakhutdinov, Urtasun, Torralba,
  and Fidler}]{DBLP:conf/iccv/ZhuKZSUTF15}
Yukun Zhu, Ryan Kiros, Richard~S. Zemel, Ruslan Salakhutdinov, Raquel Urtasun,
  Antonio Torralba, and Sanja Fidler. 2015.
\newblock \href {https://doi.org/10.1109/ICCV.2015.11} {Aligning books and
  movies: Towards story-like visual explanations by watching movies and reading
  books}.
\newblock In \emph{2015 {IEEE} International Conference on Computer Vision,
  {ICCV} 2015, Santiago, Chile, December 7-13, 2015}, pages 19--27. {IEEE}
  Computer Society.

\end{thebibliography}
\bibliographystyle{acl_natbib}

\appendix
\clearpage
\section{Appendix}
\subsection{VA-T5: LayerNorm Analysis}

Since VA-T5 is not pretrained jointly with visual and textual representation, when we combine them for finetuning we might need to re-consider how to apply the layer normalization to concatenated representations. %
Table \ref{tab:results_va_t5_clip_layernorm} contains results of no layer normalization upon fusion (concatenation of image and text representations), layer normalization only for image features (as the T5 backbone automatically normalizes text during pretraining), and layer normalization of both text and vision upon fusion. %
The differences between different ways of applying layer normalization are minor. %
Therefore, we

\subsection{Hyperparameters}

Adafactor was chosen as the optimizer as it converged faster than Adam, while taking up lesser memory; this was necessary for fitting the VA-T5-3B model on the GPU. We trained VQA-X for longer as its convergence was slower. The batch size for every model was picked based on the validation BLEU-4 score for the generated answer and rationale. For VLP, VL-T5, and VL-BART we use hyperparameters that are reported in the respective papers.
\newpage

\begin{table}[!ht]
\resizebox{\columnwidth}{!}{  
\begin{tabular}{lcc}
\toprule
                             & \textbf{Accuracy} & \textbf{BERTscore} \\
\midrule
\texttt{No LayerNorm}                 &    72.4      &    \textbf{89.3}       \\
\texttt{LayerNorm(vision)}            &    \textbf{73.0}  &        \textbf{89.3}   \\
\texttt{LayerNorm({[}vision;text{]})} &     72.6     &      \textbf{89.3}   \\
\bottomrule
\end{tabular}
}
\caption{A comparison of different way of applying layer normalization in the \vatfive-CLIP-Base model finetuned and evaluated in the \esnlive dataset.}
\label{tab:results_va_t5_clip_layernorm}
\end{table}

\begin{table*}[t]
    \centering
    \begin{subtable}{0.5\textwidth}
    \resizebox{\textwidth}{!}{  
   \begin{tabular}{llccc}
    \toprule
    \textbf{Computing Infrastructure} & NVIDIA Tesla A100  \\ 
    \midrule
    \toprule
    \textbf{Hyperparameter} & \textbf{Assignment}  \\ 
    \midrule
    Number of epochs & 20 or 50 (VQA-X) \\
    \midrule   
    Patience & 5 \\
    \midrule
    Optimizer & Adafactor \\
    \midrule
    Learning rate &  5e-5 \\
    \midrule
    Learning rate scheduler & None \\
    \midrule
    Dropout & 0.1 \\
    \midrule
    \end{tabular}}
    \caption{Common hyperparameters. }
    \label{tab:common_hyperparams_va_t5}
    \end{subtable}\par%
    \vspace*{0.5em}
    \begin{subtable}{\textwidth}
    \resizebox{\textwidth}{!}{
    \begin{tabular}{l|ccc|ccc|ccc}
    \toprule
          \multicolumn{1}{c}{} & \multicolumn{3}{c}{\textbf{\vcr}} & \multicolumn{3}{c}{\textbf{\esnlive}}       & \multicolumn{3}{c}{\textbf{\vqax}}           \\
          \cmidrule(lr){2-4}  \cmidrule(lr){5-7}  \cmidrule(lr){8-10}
          \multicolumn{1}{l}{\textbf{Model}} & \textbf{CLIP} & \textbf{Object} & \textbf{Captions} &  \multicolumn{1}{c}{\textbf{CLIP}} & \textbf{Object} & \multicolumn{1}{c}{\textbf{Captions}} & \multicolumn{1}{c}{\textbf{CLIP}} & \textbf{Object} & \textbf{Captions} \\
    \midrule
        VA-T5-Base  & 256 & 64 & 64 & 512 & 128 & 128 & 8 & 8 & 32             \\
        VA-T5-Large & 64 & - & - & 128 & - & - & 32 & - & -            \\
        VA-T5-3B    & 4 & - & - & 8 & - & - & 8 & - & -   \\
    \arrayrulecolor{black}\bottomrule
    \end{tabular}
    }
    \caption{Effective batch size. }
    \label{tab:hyperparams_va_t5_batch_size}
    \end{subtable}
    \caption{Hyperparameters for VA-T5}
    \label{tab:hyperparameters}
\end{table*}

\end{document}